\documentclass{article}
\usepackage{ijcai26}

\usepackage{times}
\usepackage{soul}
\usepackage{url}
\usepackage[hidelinks]{hyperref}
\usepackage[utf8]{inputenc}
\usepackage[small]{caption}
\usepackage{graphicx}
\usepackage{amsmath}
\usepackage{amssymb}
\usepackage{amsthm}
\usepackage{booktabs}
\usepackage{algorithm}
\usepackage{algorithmic}
\usepackage[switch]{lineno}
\usepackage{multirow}
\usepackage{booktabs}
\usepackage{tabularx}
\usepackage{placeins}
\usepackage{colortbl} 
\usepackage{arydshln}
\usepackage{pifont}
\usepackage{epsfig}
\usepackage{tcolorbox}
\tcbuselibrary{listings,breakable}
\usepackage{bm}

\usepackage[table]{xcolor}
\definecolor{red}{RGB}{166, 25, 46}  
\definecolor{blue}{RGB}{0, 70, 140}
\definecolor{green}{RGB}{43, 127, 62}

\title{QualiRAG: Retrieval-Augmented Generation for Visual Quality Understanding}

\author{
    Linhan Cao$^{1}$\footnote{Equal contribution. $^{\heartsuit}$Project lead. $^{\dag}$Corresponding authors.},\ 
    Wei Sun$^{2\ast \heartsuit }$ ,\ 
    Weixia Zhang$^{1}$, \
    Xiangyang Zhu$^{3}$, \
    Kaiwei Zhang$^{1}$,\
    Jun Jia$^{1}$, \\
    Dandan Zhu$^{2}$, \
    Guangtao Zhai$^{1}$,\ 
    Xiongkuo Min$^{1\dagger}$
    \affiliations
    \textsuperscript{\rm 1}Shanghai Jiao Tong University\\
    \textsuperscript{\rm 2}East China Normal University\\
    \textsuperscript{\rm 3}Shanghai Artificial Intelligence Laboratory \\
}

\begin{document}

\maketitle

\begin{abstract}
Visual quality assessment (VQA) is increasingly shifting from scalar score prediction toward interpretable quality understanding---a paradigm that demands \textit{fine-grained spatiotemporal perception} and \textit{auxiliary contextual information}. Current approaches rely on supervised fine-tuning or reinforcement learning on curated instruction datasets, which involve labor-intensive annotation and are prone to dataset-specific biases. To address these challenges, we propose \textbf{QualiRAG}, a \textit{training-free} \textbf{R}etrieval-\textbf{A}ugmented \textbf{G}eneration \textbf{(RAG)} framework that systematically leverages the latent perceptual knowledge of large multimodal models (LMMs) for visual quality perception. Unlike conventional RAG that retrieves from static corpora, QualiRAG dynamically generates auxiliary knowledge by decomposing questions into structured requests and constructing four complementary knowledge sources: \textit{visual metadata}, \textit{subject localization}, \textit{global quality summaries}, and \textit{local quality descriptions}, followed by relevance-aware retrieval for evidence-grounded reasoning. Extensive experiments show that QualiRAG achieves substantial improvements over open-source general-purpose LMMs and VQA-finetuned LMMs on visual quality understanding tasks, and delivers competitive performance on visual quality comparison tasks, demonstrating robust quality assessment capabilities without any task-specific training. The code will be publicly available at \url{https://github.com/clh124/QualiRAG}.
\end{abstract}

\begin{figure}[t]
\centering
\centerline{\epsfig{figure=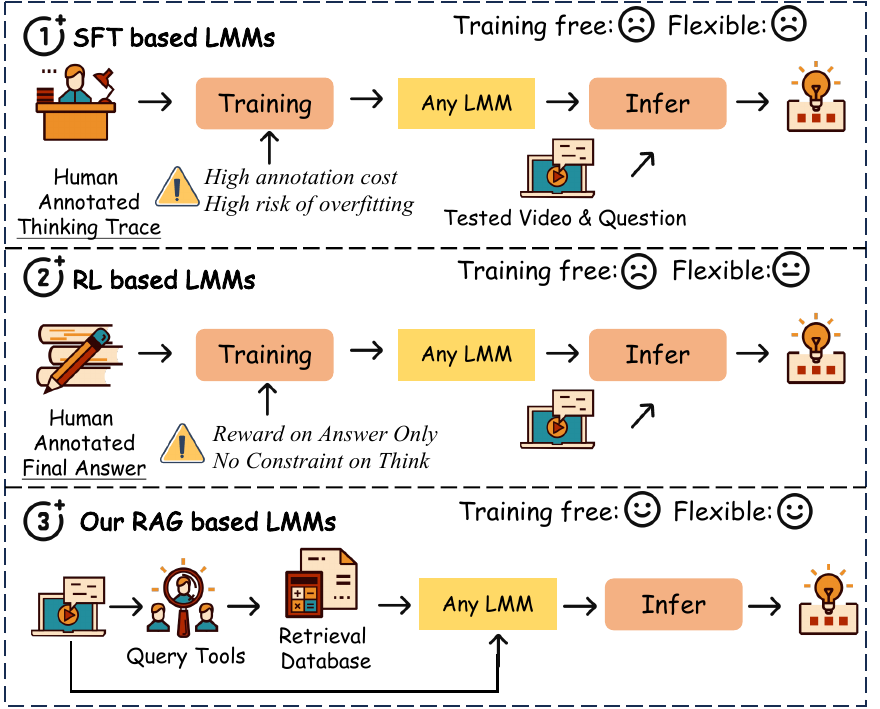,width=8.5cm}}
\caption{Comparison of training-based and training-free paradigms for visual quality understanding.
Unlike SFT- and RL-based LMMs that require task-specific training with human annotations or reward signals, our RAG-based LMM performs inference-time retrieval and reasoning without any additional training, providing improved flexibility and scalability.}
\label{fig:motivation}
\vspace{-0.4cm}
\end{figure}

\section{Introduction}

Visual quality assessment (VQA)~\cite{min2024perceptual} is a fundamental component of modern multimedia systems, serving as the primary perceptual criterion for evaluating visual content across the processing pipeline and guiding downstream decisions. Traditionally, VQA has focused on \emph{predicting scalar quality scores}~\cite{sun2022deep,wu2023exploring,wu2023q,sun2023blind,sun2025enhancing}. While such paradigms provide a compact summary of perceptual quality, they are inherently limited in interpretability: they neither explain \emph{why} an input is judged to be of high or low quality, nor identify the underlying distortion attributes. As multimedia applications increasingly demand trustworthy and explainable quality evaluation, VQA is undergoing a shift from scalar score prediction toward interpretable quality understanding.

Recently, large multimodal models (LMMs) have shown remarkable advances in visual perception and multimodal reasoning~\cite{li2024llava,bai2025qwen2,zhu2025internvl3,wang2025internvl3,yang2025qwen3}, enabling them to capture and articulate complex perceptual attributes that support visual quality understanding. To elicit such capabilities, previous studies have explored supervised fine-tuning (SFT) and reinforcement learning (RL) strategies to produce explainable quality assessment, as illustrated in Figure~\ref{fig:motivation}. Specifically, SFT-based approaches~\cite{wu2024q,jia2025scaling} depend heavily on the curation of large-scale quality instruction datasets, such as Q-Pathway~\cite{wu2024q} and OmniVQAChat-400K~\cite{jia2025scaling}. These datasets typically require a labor-intensive pipeline involving proprietary model generation and human refinement, making them expensive to scale and prone to annotation and content biases. Alternatively, RL-based approaches~\cite{li2025q,cao2025vqathinker} attempt to bypass explicit instruction data by using quality scores as implicit supervision. However, this often leads to misalignment between the generated rationale and the final prediction, resulting in inconsistent explanations. Crucially, such training-based paradigms risk compromising general-purpose reasoning and often struggle to generalize across unseen distortions and domains.

In contrast to training-based paradigms, visual quality understanding can be achieved by carefully organizing inference-time visual observations and auxiliary contextual cues. Existing LMMs~\cite{wang2025internvl3,yang2025qwen3} already encode rich knowledge related to visual distortions, temporal consistency, subject-dependent quality attributes, etc. The key challenge thus lies in how to effectively elicit and ground this knowledge with respect to the visual content under assessment. This insight naturally leads to a retrieval-augmented~\cite{lewis2020retrieval} formulation that coordinates visual inputs with auxiliary contextual retrieval, enabling fine-grained and interpretable quality reasoning without any task-specific training.

Therefore, we propose \textbf{QualiRAG}, a \textbf{training-free agentic framework} that leverages \textbf{R}etrieval-\textbf{A}ugmented \textbf{G}eneration \textbf{(RAG)} for visual quality understanding. Given a visual input and a quality-related question, QualiRAG adopts a retrieval-augmented generation pipeline composed of four collaborative modules: (1) \textbf{\textit{Query Organizer}} that reformulates the input question into structured, retrieval-oriented requests specifying subject, quality dimension, scope, and visual focus; (2) \textbf{\textit{Source Augmenter}} that constructs and populates multiple auxiliary quality knowledge databases, including \textit{visual metadata}, \textit{subject localization}, \textit{global quality summaries}, and \textit{local quality descriptions}, to produce candidate evidence; (3) \textbf{\textit{Source Selector}} that retrieves knowledge semantically aligned with the original question via relevance-aware retrieval; and (4) \textbf{\textit{Answer Generator}} that integrates the visual input with the retrieved auxiliary knowledge to generate grounded reasoning and answers. These modules decompose visual quality understanding into sequential stages of visual perception, knowledge augmentation, and evidence-grounded reasoning, enabling interpretable and fine-grained quality analysis. Extensive experiments show that QualiRAG achieves substantial improvements over open-source general-purpose LMMs and VQA-finetuned LMMs on quality understanding tasks, and delivers competitive performance on quality comparison tasks, demonstrating its effectiveness.

Our main contributions are summarized as follows:
\begin{itemize}
    \item We present \textbf{QualiRAG}, a training-free retrieval-augmented generation framework for visual quality understanding that systematically activates the latent perceptual knowledge of LMMs, without requiring any task-specific fine-tuning.
    
    \item We propose a \textbf{granularity-aware quality reasoning paradigm} that organizes visual evidence at both global and local levels, enabling interpretable inference over holistic quality perception and fine-grained, subject-dependent distortions during inference.
    
    \item We instantiate this reasoning paradigm through \textbf{four complementary auxiliary quality knowledge sources}: visual metadata, subject localization, global quality summaries, and local quality descriptions, and integrate them through relevance-aware retrieval to support evidence-grounded reasoning.
    
\end{itemize}

\section{Related Work}

\noindent\textbf{LMMs for Visual Quality Assessment.}

Recent advances in LMMs have sparked growing interest in their application to visual quality assessment. Existing \textit{training-based} methods can be broadly categorized into three paradigms. The first paradigm~\cite{wang2025aigv,liu2025improving} leverages LMMs as powerful feature extractors, where hidden states from LMM backbones are fed into lightweight prediction heads to regress a scalar quality score. Although benefiting from the powerful feature representations of LMMs, these methods still adhere to a conventional score-regression paradigm and fail to fully exploit the generative and reasoning capabilities of LMMs. The second paradigm~\cite{jia2024vqa,jia2025scaling} formulates VQA as a vision-language instruction task. By constructing visual instruction–answer pairs, these methods fine-tune LMMs via supervised learning to generate quality-related descriptions. While this paradigm enhances interpretability and better aligns model outputs with human perceptual judgments, the collection of large-scale, high-quality instruction datasets is expensive and difficult to scale. The third paradigm~\cite{li2025q,cao2025vqathinker} introduces reinforcement learning into VQA, typically optimized via Group Relative Preference Optimization (GRPO)~\cite{shao2024deepseekmath}. In this setting, LMMs are trained to generate step-by-step reasoning before predicting a quality score. The reward signal is computed based on the final predicted score, while the intermediate reasoning process remains unsupervised. This may lead to reasoning traces that are inconsistent with or even contradictory to the final quality prediction.

Beyond these training-based paradigms, several studies have explored \textit{training-free} approaches for VQA. In-context learning~\cite{chen2023x,wu2024comprehensive} exploits carefully designed prompts or a small number of exemplars to elicit the quality perception capabilities of LMMs. Other works~\cite{zhu2025agenticiqa,xing2025q} adopt agent-based frameworks for visual quality perception, where fine-tuned expert models or proprietary LMMs are employed as specialized perception modules to enhance quality assessment ability. Our QualiRAG follows the agent-based paradigm but differs fundamentally in that it leverages RAG to systematically exploit the capabilities of LMMs, without relying on additional quality-labeled data for training or task-specific VQA models as expert modules.

\begin{figure*}[t]
\centering
\centerline{\epsfig{figure=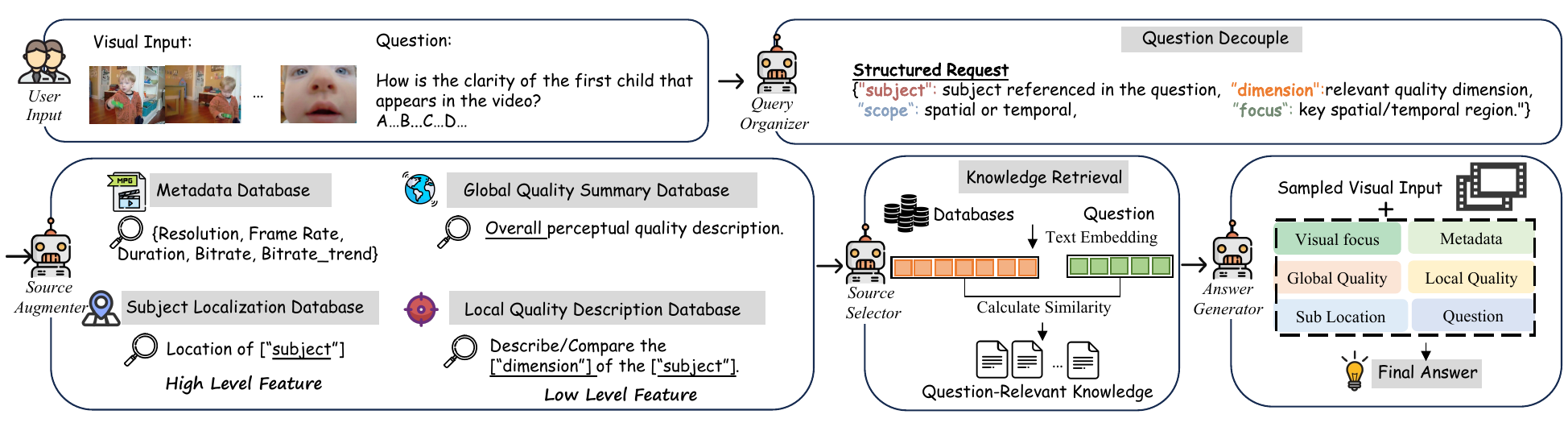,width=18cm}}
\vspace{-0.2cm}
\caption{The overall workflow of our QualiRAG. Given a visual input and a quality-related question, QualiRAG first decomposes the query into structured requests via a \textbf{Query Organizer}, then constructs complementary auxiliary knowledge sources through a \textbf{Source Augmenter}. A \textbf{Source Selector} retrieves question-relevant evidence from multiple knowledge databases, which is finally integrated with sampled visual inputs by the \textbf{Answer Generator} to produce evidence-grounded and interpretable quality reasoning.}
\label{fig:model_framework}
\vspace{-0.4cm}
\end{figure*}

\vspace{0.1cm}

\noindent\textbf{Retrieval-Augmented Generation for LMMs.} RAG has emerged as an effective paradigm for enhancing LMMs by grounding generation on externally retrieved knowledge. By incorporating relevant information from curated databases, RAG enables LMMs to produce responses that are more factual, interpretable, and domain-aware, while mitigating hallucination~\cite{gao2023retrieval}. A typical RAG pipeline consists of three stages: (1) preprocessing and organizing raw data into a structured knowledge base; (2) retrieving task-relevant information conditioned on the input query; (3) conditioning the generation process on the retrieved content. RAG has been successfully applied to a variety of multimodal tasks, including visual question answering~\cite{lin2023fine,luo2024video}, visual captioning~\cite{wu2024dir,li2024understanding}, and multimodal content generation~\cite{zhang2024garmentaligner,lyu2025realrag}. In this work, we extend RAG to visual quality understanding, demonstrating its effectiveness in grounding quality-aware multimodal reasoning.

\section{Method}
\label{sec:method}

The framework of QualiRAG is illustrated in Figure~\ref{fig:model_framework}. Given a visual input $V$ and a quality-related question $Q$, QualiRAG adopts a training-free, retrieval-augmented reasoning pipeline composed of four collaborative modules: (1) \textbf{Query Organizer} that reformulates the input question into structured, retrieval-oriented requests; (2) \textbf{Source Augmenter} that constructs and populates multiple auxiliary quality knowledge databases to produce candidate evidence; (3) \textbf{Source Selector} that retrieves knowledge that is semantically aligned with the original question; and (4) \textbf{Answer Generator} that integrates visual inputs with the retrieved auxiliary knowledge to generate grounded reasoning and the final answer. 
In practice, the framework is instantiated using two LMMs, denoted as $\mathrm{LMM}_{\text{main}}$ and $\mathrm{LMM}_{\text{aux}}$. The main LMM, $\mathrm{LMM}_{\text{main}}$, performs query interpretation and final reasoning, while the auxiliary LMM, $\mathrm{LMM}_{\text{aux}}$, is responsible for auxiliary knowledge construction.

\subsection{Query Organizer: Granularity-Aware Query Formulation}

The Query Organizer reformulates quality-related questions into structured representations that shape the granularity of subsequent perception and retrieval. Given a quality-related question $Q$, we perform structured question decoupling to extract its key semantic components in a text-only manner, where a LMM is prompted to analyze the question without accessing any visual inputs and generate a structured request that specifies what auxiliary information should be queried for quality reasoning.

Concretely, we prompt the LMM with a decoupling instruction $\mathcal{P}_{\text{dec}}$ (see Appendix B.1) to transform the original question $Q$ into a structured request $\mathbf{R}$ as follows:
\begin{equation}
\mathbf{R} = \mathrm{LMM}_{\text{main}}(\mathcal{P}_{\text{dec}} , Q),
\end{equation}
where $\mathbf{R}$ is represented as a JSON object with four fields:
\begin{equation}
\mathbf{R} =
\left\{
\texttt{subject},\ \texttt{dimension}, \texttt{scope}, \texttt{focus}
\right\}.
\end{equation}
Here, \texttt{subject} identifies the primary visual entity referenced in the question, \texttt{dimension} specifies the relevant quality dimension (\textit{e.g.}, blur, noise, artifact, or temporal stability), \texttt{scope} indicates whether the queried quality concerns spatial or temporal aspects, and \texttt{focus} denotes the spatial region or temporal segment emphasized by the question. If a component is not explicitly mentioned or cannot be reliably inferred from the question, the corresponding field is set to \texttt{NULL}.

\subsection{Source Augmenter: Multi-Source Knowledge Generation}
The Source Augmenter construct multiple auxiliary knowledge databases, denoted as $\mathcal{K}$, which provide query-specific, multi-granularity contextual information beyond raw visual input. Specifically, we build four complementary knowledge databases: (1) a metadata database $\mathcal{K}_{\text{meta}}$, (2) a subject localization database $\mathcal{K}_{\text{loc}}$, (3) a global quality summary database $\mathcal{K}_{\text{globalQ}}$, and (4) a local quality description database $\mathcal{K}_{\text{localQ}}$.
\vspace{0.1cm}

\noindent\textbf{Metadata Database.}
Existing LMMs infer visual quality primarily from perceptual cues in the visual content, without explicitly observing low-level technical properties, such as codec- and bitrate-level signals that characterize the underlying acquisition and compression process. Such technical factors provide important context for interpreting compression artifacts and temporal variations in visual quality. To make these signals available during reasoning, we extract a set of technical metadata using \texttt{ffprobe}~\cite{ffprobe}, including spatial resolution $r$, frame rate $f$, video duration $d$, average bitrate $b_{\text{avg}}$, and temporal bitrate variation trends $b_{\text{trend}}$. For image inputs, only the spatial resolution is available and thus included as metadata. Formally, the metadata database is defined as
\begin{equation}
\mathcal{K}_{\text{meta}}(V) = \{ r, f, d, b_{\text{avg}}, b_{\text{trend}} \}.
\end{equation}

\vspace{0.1cm}
\noindent\textbf{Subject Localization Database.}
Although modern LMMs demonstrate strong object recognition capabilities, their spatial grounding is often coarse, which limits their effectiveness for queries involving \textit{local quality distortions} that require accurate region-level assessment. To provide explicit spatial grounding, we construct a subject localization database that identifies the spatial extent of the queried subject across sampled frames.  When a subject is specified in $\mathbf{R}$, we localize it in each frame of $\mathcal{F}s$ using APE~\cite{shen2024aligning}, a prompt-driven object detection model that takes natural language descriptions as input and returns subject-specific bounding regions. The subject localization database is defined as

\begin{equation}
\mathcal{K}_{\text{loc}}(V, s)
=
\left\{
(t,\mathcal{B}_t)\ \middle|\ t \in \mathcal{F}_s
\right\},
\end{equation}
where $s = \mathbf{R}[\texttt{subject}]$, and $\mathcal{B}_t$ denotes the predicted bounding region of the queried subject in sampled frame $t$.

\vspace{0.1cm}
\noindent\textbf{Global Quality Summary Database.}
The global quality summary database captures the overall perceptual characteristics of a visual input at a coarse level. For each image or video, we prompt the queried LMM with a unified global quality query $Q_{\text{global}}$ (see Appendix B.2) to generate a holistic description that reflects coarse-grained quality attributes, such as overall clarity, stability, color fidelity, and compression artifacts at the image or video level. Formally, the global quality summary database is defined as
\begin{equation}
\mathcal{K}_{\text{globalQ}}(V) = \mathrm{LMM}_{\text{aux}}(V, Q_{\text{global}}).
\end{equation}

\begin{table*}[t]\small
    \centering
    \renewcommand\arraystretch{1.1}
    \renewcommand\tabcolsep{14pt}
    \belowrulesep=0pt\aboverulesep=0pt
    \caption{Results on the {\tt dev} subset of Q-bench for the image quality understanding ability of LMMs. The \textbf{Labels} column indicates whether a model requires VQA-lablled data for training. Models shown in gray text correspond to closed-sourced proprietary LMMs. \textbf{\textcolor{red}{Red}}, \textbf{\textcolor{blue}{blue}}, and \textbf{bold} mark the top three results after excluding proprietary models, consistent with the following tables.}
     \vspace{-8pt}
    \resizebox{\linewidth}{!}{\begin{tabular}{l|c|ccc|cc|cc|c}
    \toprule
        \multicolumn{2}{c|}{\textbf{Sub-categories}} & \multicolumn{3}{c|}{\textbf{Question Types}} & \multicolumn{4}{c|}{\textbf{Quality Concerns}} & \multirow{2}{*}{{\textit{Overall$\uparrow$}}} \\ \cdashline{1-9}
        \textbf{LMM} & \textbf{Labels}
        & {\textit{Yes-or-No$\uparrow$}}
        & {\textit{What$\uparrow$}} 
        & {\textit{How$\uparrow$}} 
        & \textit{Distortion$\uparrow$} 
        & \textit{Other$\uparrow$} 
        & \textit{I-C Distortion$\uparrow$}  
        & \textit{I-C Other$\uparrow$} \\ \hline

        \rowcolor{gray!10}
        \multicolumn{10}{l}{\textit{Open-sourced General-purpose LMMs $\sim$ 8B}} \\ 
        LLaVA-OneVision-7B & \ding{55}
        & 81.27\% & 80.31\% & 68.56\% & 73.74\% & \textbf{\textcolor{blue}{77.78\%}} & 74.67\% & 84.08\% & 76.79\% \\

         mPLUG-Owl3-7B & \ding{55} & 63.48\% & 77.85\% & 50.87\% & 57.80\% & 66.89\% & 67.33\% & 64.00\% & 63.40\% \\

        InternVL3-9B & \ding{55}
        & 80.18\% & 78.76\% & 68.76\% & 72.76\% & 75.69\% & 73.68\% & \textbf{\textcolor{blue}{86.12\%}} & 75.99\%  \\

        InternVL3.5-8B & \ding{55}
        & 80.36\% & 79.87\% & 69.78\% & 75.10\% & 74.54\% & 76.64\% & 84.08\% & 76.72\% \\

        Qwen2.5-VL-7B & \ding{55}
        & 81.45\% & \textbf{83.19\%} & 68.97\% & 75.10\% & 76.85\% & 77.96\% & \textbf{85.31\%} & 77.86\% \\

        Qwen3-VL-8B & \ding{55}
        & 80.91\% & \textbf{\textcolor{blue}{83.85\%}} & \textbf{70.99\%} & \textbf{\textcolor{blue}{77.43\%}} & \textbf{77.08\%} & \textbf{78.62\%} & 83.27\% & \textbf{78.53\%} \\
        
        GLM-4.1V-9B & \ding{55} & 81.45\% & 80.53\% & 69.98\% & 76.07\% & 75.23\% & 76.64\% & 84.90\% & 77.39\%  \\

        \hline
        \rowcolor{gray!10}
        \multicolumn{10}{l}{\textit{VQA-finetuned LMMs}} \\ 
        Q-Instruct & \ding{52}
        & 76.91\% & 65.04\% & 55.78\% & 64.01\% & 67.13\% & 64.80\% & 71.84\% & 66.35\%   \\

        Co-Instruct & \ding{52}
        & \textbf{82.00\%} & 76.77\% & 66.73\% & 75.29\% & 71.53\% & 74.67\% & 83.27\% & 75.38\%   \\

        Q-Insight & \ding{52}
        & \textbf{82.00\%} & 82.30\% & 68.56\% & 76.07\% & 76.39\% & 77.30\% & 83.67\% & 77.66\%  \\

        VisualQuality-R1 
        & \ding{52}
        & \textbf{\textcolor{blue}{82.91\%}} 
        & 82.30\% & \textbf{\textcolor{blue}{71.60\%}} 
        & \textbf{\textcolor{red}{77.63\%}} 
        & \textbf{\textcolor{blue}{77.78\%}}
        & \textbf{\textcolor{blue}{79.61\%}}
        & 83.27\% 
        & \textbf{\textcolor{blue}{79.00\%}} \\

        \hline
        \rowcolor{gray!10}
        \multicolumn{10}{l}{\textit{Closed-sourced Proprietary LMMs}} \\

        {\color{black!60}GPT-5} & {\color{black!60}\ding{55}}
        & {\color{black!60}82.18\%} 
        & {\color{black!60}84.51\%}  
        & {\color{black!60}74.24\%} 
        & {\color{black!60}77.63\%}
        & {\color{black!60}79.86\%} 
        & {\color{black!60}80.92\%}
        & {\color{black!60}85.71\%}
        & {\color{black!60}80.27\%}  \\

        {\color{black!60}Gemini-3-Pro} & {\color{black!60}\ding{55}}
        & {\color{black!60}82.91\%} 
        & {\color{black!60}88.48\%}
        & {\color{black!60}71.05\%} 
        & {\color{black!60}77.33\%}
        & {\color{black!60}78.83\%} 
        & {\color{black!60}83.80\%}
        & {\color{black!60}86.67\%}
        & {\color{black!60}80.63\%} \\

        \hline
        \rowcolor{gray!10}
        \multicolumn{10}{l}{\textit{Our RAG-based LMM}} \\ 
        QualiRAG 
        & \ding{55}
        & \textbf{\textcolor{red}{85.82\%}} 
        & \textbf{\textcolor{red}{84.60\%}} 
        & \textbf{\textcolor{red}{74.59\%}} 
        & \textbf{77.01\%} 
        & \textbf{\textcolor{red}{81.71\%}} 
        & \textbf{\textcolor{red}{85.20\%}} 
        & \textbf{\textcolor{red}{87.35\%}} 
        & \textbf{\textcolor{red}{81.74\%}}  \\

       \bottomrule
    \end{tabular}}
    \vspace{-5pt}
    \label{tab:q-bench-dev}
\end{table*}

\begin{table*}[t]\small
    \centering
    \renewcommand\arraystretch{1.1}
    \renewcommand\tabcolsep{14pt}
    \belowrulesep=0pt\aboverulesep=0pt
    \caption{Results on the {\tt dev} subset of Q-Bench-Video for the video quality understanding ability of LMMs. Q-Router (GPT-4o) indicates that GPT-4o is used as the reasoning backbone. Note that Q-Router integrates multiple label-trained no-reference VQA models and is therefore categorized as a training-based method.}
    \vspace{-8pt}
    \resizebox{\linewidth}{!}{\begin{tabular}{l|c|ccc|cc|cc|c}
    \toprule
        \multicolumn{2}{c|}{\textbf{Sub-categories}} & \multicolumn{3}{c|}{\textbf{Question Types}} & \multicolumn{4}{c|}{\textbf{Quality Concerns}} & \multirow{2}{*}{{\textit{Overall$\uparrow$}}} \\ \cdashline{1-9}
        \textbf{LMM} & \textbf{Labels}
        & {\textit{Yes-or-No$\uparrow$}}
        & {\textit{What-How$\uparrow$}} 
        & {\textit{Open-ended$\uparrow$}} 
        & \textit{Tech.$\uparrow$} 
        & \textit{Aes.$\uparrow$} 
        & \textit{Temp.$\uparrow$}  
        & \textit{AIGC}$\uparrow$ \\ 
    \hline
    \rowcolor{gray!10}
        \multicolumn{10}{l}{\textit{Open-sourced General-purpose LMMs $\sim$ 8B}} \\ 
        LLaVA-OneVision-7B & \ding{55}
        & 62.13\% & 52.23\% & 38.56\% & 48.74\% & 61.53\% & 48.81\% & \textbf{44.57\%} & 52.12\% \\
         mPLUG-Owl3-7B & \ding{55} & 60.82\% & 56.52\% & 35.84\% & 51.34\% & 60.46\% & 54.26\% & 37.30\% & 52.44\%\\
        InternVL3-9B & \ding{55}
        & 56.57\% & 51.58\% & 30.40\% & 46.35\% & 59.80\% & 50.83\% & 31.51\% & 47.61\% \\
        InternVL3.5-8B & \ding{55}
        & 61.97\% & 51.13\% & 35.31\% & 49.28\% & \textbf{65.17\%} & 54.33\% & 29.14\% & 50.70\% \\
        Qwen2.5-VL-7B & \ding{55}
        & 68.54\% & \textbf{57.01\%} & 35.56\% & 54.94\% & 63.12\% & 57.27\% & 35.70\% & 55.30\% \\
        Qwen3-VL-8B & \ding{55}
        & 62.44\% & 51.81\% & 38.92\% & 51.68\% & 62.85\% & 54.90\% & 31.94\% & 52.11\% \\
        GLM-4.1V-9B & \ding{55} & 63.85\% & 47.96\% & 31.45\% & 49.28\% & 60.77\% & 50.59\% & 24.25\% & 49.15\% \\
    \hline
    \rowcolor{gray!10}
        \multicolumn{10}{l}{\textit{VQA-finetuned LMMs}} \\ 
        VQA$^2$ & \ding{52}
        & \textbf{73.81\%} & 56.40\% & 38.33\% & \textbf{60.70\%} & 56.65\% & 61.09\% & 38.11\% & 56.67\% \\
        VQAThinker & \ding{52}
        & 70.42\% & 56.56\% & 35.15\% & 55.95\% & 64.56\% & \textbf{\textcolor{blue}{61.69\%}} & 30.91\% & 55.70\% \\
        OmniVQA-Chatter & \ding{52}
        & \textbf{\textcolor{blue}{75.51\%}} & \textbf{\textcolor{blue}{59.76\%}} & \textbf{40.37\%} & \textbf{\textcolor{red}{62.05\%}} & 61.58\% & \textbf{\textcolor{red}{63.45\%}} & 42.38\% & \textbf{59.08\%} \\
        
        Q-Router (GPT-4o)
        & \ding{52}
        & \textbf{\textcolor{red}{76.00\%}} 
        & \textbf{57.01\%} 
        & \textbf{\textcolor{blue}{43.33\%}} 
        & 59.31\% 
        & \textbf{\textcolor{blue}{65.31\%}} 
        & \textbf{61.46\%} 
        & \textbf{\textcolor{red}{50.22\%}} 
        & \textbf{\textcolor{blue}{60.07\%}} \\  
    \hline
    \rowcolor{gray!10}
        \multicolumn{10}{l}{\textit{Closed-sourced Proprietary LMMs}} \\ \hdashline

        {\color{black!60}GPT-5} 
        & {\color{black!60}\ding{55}}
        & {\color{black!60}68.72\%} 
        & {\color{black!60}56.49\%} 
        & {\color{black!60}43.21\%} 
        & {\color{black!60}56.58\%} 
        & {\color{black!60}68.99\%} 
        & {\color{black!60}59.91\%} 
        & {\color{black!60}37.42\%} 
        & {\color{black!60}57.22\%} \\

        {\color{black!60}Gemini-3-Pro}
        & {\color{black!60}\ding{55}}
        & {\color{black!60}76.78\%}
        & {\color{black!60}61.00\%} 
        & {\color{black!60}47.87\%}
        & {\color{black!60}61.00\%} 
        & {\color{black!60}69.56\%} 
        & {\color{black!60}59.49\%} 
        & {\color{black!60}60.00\%} 
        & {\color{black!60}63.02\%} \\
    \hline
    \rowcolor{gray!10}
        \multicolumn{10}{l}{\textit{Our RAG-based LMM}} \\ 
        QualiRAG & \ding{55}
        & 73.18\%
        & \textbf{\textcolor{red}{62.95\%}} 
        & \textbf{\textcolor{red}{44.95\%}} 
        & \textbf{\textcolor{blue}{61.34\%}} 
        & \textbf{\textcolor{red}{70.74\%}} 
        & \textbf{\textcolor{blue}{61.69\%}} 
        & \textbf{\textcolor{blue}{48.17\%}}
        & \textbf{\textcolor{red}{61.66\%}} \\
    \bottomrule
    \end{tabular}}
    \vspace{-8pt}
    \label{tab:q-bench-video-dev}
\end{table*}

\vspace{0.1cm}
\noindent\textbf{Local Quality Description Database.}
To complement global perceptual summaries, the local quality description database captures on fine-grained, question-driven, and subject-aware quality attributes. The local quality query $Q_{\text{local}}$ (see Appendix B.2) is instantiated based on the \texttt{subject} and \texttt{dimension} fields in $\mathbf{R}$. When $\mathbf{R}$ corresponds to a single visual input, $Q_{\text{local}}$ prompts the LMM to describe the specified quality dimension of the queried subject; for multiple visual inputs, it is adapted to elicit comparative descriptions across visual inputs. 

The specific visual input provided to the LMM is determined by the \texttt{scope} and \texttt{focus} fields in $\mathbf{R}$: for spatial-scope queries, subject-centric regions corresponding to the queried \texttt{subject} are extracted using the subject localization database $\mathcal{K}_{\text{loc}}$, optionally constrained by the specified \texttt{focus}, to support region-level quality analysis; for temporal-scope queries, full video frames are retained to preserve the temporal segment indicated by \texttt{focus} and capture quality variations over time.

To consolidate local quality information, the LMM is queried $n_l$ times, and consistent content across the generated descriptions is aggregated to form the local quality description database $\mathcal{K}_{\text{localQ}}$:
\begin{equation}
\label{local}
\mathcal{K}_{\text{localQ}}(V, \mathbf{R}) 
= \mathrm{Aggregate}\big(\{\mathrm{LMM}_\text{aux}(V, Q_{\text{local}}(\mathbf{R}))\}_{i=1}^{n_l}\big),
\end{equation}
where detailed implementation of the aggregation strategy is provided in Appendix B.2.

\subsection{Source Selector: Relevance-Aware Knowledge Retrieval}

The Source Selector aims to retrieve evidence that is semantically aligned with the original quality-related question from auxiliary knowledge databases $\mathcal{K}$, while suppressing redundant or irrelevant information. This module serves two purposes: (1) filtering noisy or weakly related knowledge candidates, and (2) reducing the overall context length to improve inference efficiency and reasoning focus.

Following Video-RAG~\cite{luo2024video}, we encode the input question $Q$ and each candidate knowledge entry $d \in \mathcal{K}$ into a shared semantic embedding space using the \textbf{Contriever} framework~\cite{izacard2021unsupervised}. Semantic relevance between the question and each knowledge unit is computed via dense inner-product similarity. To enable scalable and efficient retrieval, we index all knowledge embeddings using the \textbf{FAISS} library~\cite{johnson2019billion} and perform approximate nearest-neighbor search. Only knowledge entries whose similarity scores exceed a predefined threshold $\tau$ are retained for subsequent reasoning. Formally, the filtered knowledge set is defined as:
\begin{equation}
\label{eq:sim}
\mathcal{K}^{*}(Q) =
\left\{ d \in \mathcal{K} \;\middle|\;
\mathrm{sim}(d, Q) \ge \tau
\right\},
\end{equation}
where $\mathrm{sim}(\cdot,\cdot)$ denotes the dense inner-product similarity computed via inner-product search.

\subsection{Answer Generator: Knowledge-Grounded Quality Reasoning}

The Answer Generator constitutes the final stage of QualiRAG, where the visual input $V$, the quality-related question $Q$, and the retrieved auxiliary knowledge $\mathcal{K}^{*}$ are jointly integrated to produce grounded quality reasoning and answers. By conditioning the LMM on these complementary inputs, the model aligns perceptual observations with external quality-related evidence, thereby mitigating hallucination and improving reasoning fidelity. The final answer $A$ is generated as
\begin{equation}
A = \mathrm{LMM}_{\text{main}}(V, \mathcal{K}^{*}, Q).
\end{equation}

\begin{table}[t]\small
    \centering
    \renewcommand\arraystretch{1.1}
    \renewcommand\tabcolsep{7pt}
    \belowrulesep=0pt\aboverulesep=0pt
    \caption{Performance comparison on the image quality comparison task. $\Diamond$ denotes methods trained specifically for quality scoring, while $\spadesuit$ indicates question-answering–based methods. These conventions are consistent with those in Table~\ref{tab:video_quality}.}
    \vspace{-8pt}
    \resizebox{\linewidth}{!}{\begin{tabular}{l|c|ccc|c}
    \toprule
        \multicolumn{2}{c|}{\textbf{Benchmarks}} & \multirow{2}{*}{\textbf{LIVE-C}}  & \multirow{2}{*}{\textbf{AGIQA}}  & \multirow{2}{*}{\textbf{PIPAL}} & \multirow{2}{*}{{\textit{Overall$\uparrow$}}} \\ \cdashline{1-2}
        \textbf{Models}  & \textbf{Labels}  & & & & \\ \hline
        \rowcolor{gray!10}
        \multicolumn{6}{l}{\textit{Open-sourced General-purpose LMMs $\sim$ 8B}} \\ 
        $\spadesuit$ LLaVA-OneVision-7B & \ding{55} & 76.54\% & \textbf{78.95\%} & 67.20\% & 76.12\%\\
        $\spadesuit$ InternVL3-9B & \ding{55} & 66.10\% & 64.05\% & 57.00\% & 63.15\%\\
        $\spadesuit$ InternVL3.5-8B & \ding{55} & 69.35\% & 64.86\% & 58.80\% & 64.70\% \\
        $\spadesuit$ Qwen2.5-VL-7B & \ding{55} & 78.42\% & 78.40\% & \textbf{\textcolor{blue}{70.40\%}} & 76.85\% \\
        $\spadesuit$ Qwen3-VL-8B & \ding{55} & 81.67\% & 77.19\% & 68.40\% & 76.50\% \\

        \hline
        \rowcolor{gray!10}
        \multicolumn{6}{l}{\textit{DNN-based VQA Methods}} \\
        $\Diamond$ MUSIQ &  \ding{52} & 82.36\% & 70.82\% & 66.80\% & 72.66\% \\
        $\Diamond$ TRES & \ding{52} & 81.85\% & 72.03\% & 68.00\% & 73.47\% \\
        $\Diamond$ CLIP-IQA+ &  \ding{52} & 84.18\% & 74.74\% & 65.40\% & 75.07\% \\
        $\Diamond$ LIQE &  \ding{52} & \textbf{85.96\%} & 74.60\% & 68.40\% & 75.97\%\\
        $\Diamond$ TOPIQ & \ding{52} & 82.36\% & 73.96\% & 66.40\% & 74.40\% \\
        \hline
        \rowcolor{gray!10}
        \multicolumn{6}{l}{\textit{LMM-based VQA Methods}} \\ 

        $\Diamond$ Q-Align &  \ding{52} & \textbf{\textcolor{red}{87.33\%}} & \textbf{\textcolor{blue}{79.07\%}} & 67.00\% & \textbf{\textcolor{blue}{78.60\%}} \\
        $\spadesuit$ Q-Instruct &  \ding{52} & 56.68\% & 62.91\% & 60.60\% & 61.05\% \\
        $\spadesuit$ Co-Instruct &  \ding{52} & 83.56\% & 77.20\% & 66.80\% & 76.62\% \\
        $\Diamond$ Q-Insight &  \ding{52} & 84.59\% & 76.94\% & 68.00\% & 76.94\% \\
        $\Diamond$ VisualQuality-R1 &  \ding{52} & \textbf{\textcolor{blue}{86.30\%}} & 78.29\% & \textbf{68.80\%} & \textbf{78.26\%} \\

        \hline
        \rowcolor{gray!10}
        \multicolumn{6}{l}{\textit{Ours RAG-based Method}} \\ 
         $\spadesuit$ QualiRAG & \ding{55} & 85.62\% & \textbf{\textcolor{red}{79.28\%}} & \textbf{\textcolor{red}{73.40\%}} & \textbf{\textcolor{red}{79.58\%}} \\     
       
       \bottomrule
    \end{tabular}}
    \vspace{-5pt}
    \label{tab:image_quality}
\end{table}

\begin{table}[t]\small
    \centering
    \renewcommand\arraystretch{1.1}
    \renewcommand\tabcolsep{4pt}
    \belowrulesep=0pt\aboverulesep=0pt
    \caption{Performance comparison on the video quality comparison task.}
    \vspace{-8pt}
    \resizebox{\linewidth}{!}{\begin{tabular}{l|c|ccc|c}
    \toprule
        \multicolumn{2}{c|}{\textbf{Benchmarks}} & \multirow{2}{*}{{\textbf{KoNViD-1K}}} & \multirow{2}{*}{\textbf{VDPVE}} & \multirow{2}{*}{\textbf{LIVE-HFR}} & \multirow{2}{*}{{\textit{Overall$\uparrow$}}} \\ \cdashline{1-2}
        \textbf{Models}  & \textbf{Labels}  & & & \\ \hline
        \rowcolor{gray!10}
        \multicolumn{6}{l}{\textit{Open-sourced General-purpose LMMs $\sim$ 8B}} \\ 

        $\spadesuit$ LLaVA-OneVision-7B & \ding{55} & 75.33\% & 67.30\% & \textbf{62.92\%} & 70.29\% \\
        $\spadesuit$ InternVL3-9B & \ding{55} & 71.00\% & 61.34\% & 57.92\% & 65.29\%\\
        $\spadesuit$ InternVL3.5-8B & \ding{55} & 73.33\% & 68.74\% & 58.33\% & 68.94\% \\
        $\spadesuit$ Qwen2.5-VL-7B & \ding{55} & 73.50\% & 68.50\% & 60.42\% & 69.34\% \\
        $\spadesuit$ Qwen3-VL-8B  & \ding{55} & 75.33\% & 69.69\% & 59.58\% & 70.45\% \\

        \hline
        \rowcolor{gray!10}
        \multicolumn{6}{l}{\textit{DNN-based VQA Methods}} \\ 

        $\Diamond$ FAST-VQA & \ding{52} & 84.17\% & 72.79\% & 60.42\% & 75.86\%\\
        $\Diamond$ DOVER & \ding{52} & 85.50\% & 73.51\% & 62.50\% & 77.13\% \\
        $\Diamond$ COVER & \ding{52} & 84.17\% & 73.03\% & \textbf{\textcolor{blue}{63.33\%}} & 76.49\%  \\
        $\Diamond$ MinimalisticVQA & \ding{52} & 83.33\% & 72.79\% & 56.67\% & 74.74\% \\ 

        \hline
        \rowcolor{gray!10}
        \multicolumn{6}{l}{\textit{LMM-based VQA Methods}} \\ 
        $\Diamond$ Q-Align & \ding{52} & \textbf{86.17\%} & 74.22\% & 62.08\% & \textbf{77.60\%} \\
        $\spadesuit$ Q-Instruct & \ding{52} & 56.50\% & 54.18\% & 49.17\% & 54.33\% \\
        $\spadesuit$ Co-Instruct & \ding{52} & 76.50\% & 65.87\% & 51.25\% & 68.15\% \\
        $\Diamond$ VQA$^2$ & \ding{52} & \textbf{\textcolor{blue}{86.50\%}} & \textbf{\textcolor{blue}{76.13\%}} & 60.00\% & \textbf{\textcolor{blue}{78.00\%}} \\
        $\Diamond$ VQAThinker & \ding{52} & \textbf{\textcolor{red}{87.17\%}} & \textbf{\textcolor{red}{77.33\%}} & 62.08\% & \textbf{\textcolor{red}{79.11\%}} \\

        \hline
        \rowcolor{gray!10}
        \multicolumn{6}{l}{\textit{Ours RAG-based Method}} \\
         $\spadesuit$ QualiRAG & \ding{55} & 80.33\% & \textbf{74.46\%} & \textbf{\textcolor{red}{63.75\%}} & 75.22\% \\     
    
       \bottomrule
    \end{tabular}}
    \vspace{-5pt}
    \label{tab:video_quality}
\end{table}

\section{Experiments}
\subsection{Experimental Setups}

We evaluate QualiRAG from two complementary perspectives: \textbf{visual quality understanding} and \textbf{visual quality comparison}. The former constitutes the primary task of our model, while the latter serves as a dedicated evaluation protocol for visual quality rating, another important visual quality perception task, enabling direct comparison with models specifically designed for quality rating.

\vspace{0.1cm}

\noindent\textbf{Validation Benchmarks.} For \textbf{visual quality understanding}, we evaluate the models on two representative benchmarks, \textbf{Q-Bench}~\cite{wu2023qbench} and \textbf{Q-Bench-Video}~\cite{zhang2025q}, which assess LMMs’ ability to perceive and reason about visual quality attributes through question answering. Q-Bench targets image-level quality understanding with tasks probing both global and localized perceptual factors, while Q-Bench-Video extends the evaluation to videos, covering technical, aesthetic, temporal, and AIGC-related distortions under both single-video and pairwise comparison settings. We report results on the publicly available {\tt dev} subsets of both benchmarks.

For \textbf{visual quality comparison}, we evaluate on three image benchmarks: LIVE-C~\cite{ghadiyaram2015massive}, AGIQA~\cite{li2023agiqa}, and PIPAL~\cite{jinjin2020pipal}, and three video benchmarks: KoNViD-1k~\cite{hosu2017konstanz}, VDPVE~\cite{gao2023vdpve}, and LIVE-HFR~\cite{madhusudana2021subjective}. Among these benchmarks, LIVE-C and KoNViD-1k focus on user-generated content captured in unconstrained real-world scenarios with authentic distortions. AGIQA evaluates image quality understanding for AI-generated content, while PIPAL and VDPVE target quality variations introduced by image and video processing pipelines, such as compression and enhancement. LIVE-HFR specifically benchmarks temporal perceptual distortions arising from different frame rates. For each benchmark, we iteratively select one image or video and randomly sample another instance to form a comparison pair. Once constructed, the same set of pairs is used for all evaluated methods to ensure fair comparison.

\vspace{0.1cm}

\noindent\textbf{Competing Methods.}
For \textbf{visual quality understanding}, we compare QualiRAG against three categories of LMMs:
\begin{itemize}
\item \textbf{Open-source general-purpose LMMs}, including LLaVA-OneVision-7B~\cite{li2024llava}, mPLUG-Owl3-7B~\cite{ye2024mplug}, InternVL3-9B~\cite{zhu2025internvl3}, InternVL3.5-8B~\cite{wang2025internvl3}, Qwen2.5-VL-7B~\cite{bai2025qwen2}, Qwen3-VL-8B~\cite{yang2025qwen3}, and GLM-4.1V-9B~\cite{hong2025glm}.
\item \textbf{VQA-finetuned LMMs}, which are specifically adapted for quality-related question answering. This group includes Q-Instruct~\cite{wu2024q}, Co-Instruct~\cite{wu2024towards}, Q-Insight~\cite{li2025q}, and VisualQuality-R1~\cite{wu2025visualquality} for image-based quality understanding, as well as VQA$^2$~\cite{jia2024vqa}, VQAThinker~\cite{cao2025vqathinker}, OmniVQA-Chatter~\cite{jia2025scaling}, and Q-Router~\cite{xing2025q} for video-based settings.
\item \textbf{Closed-sourced proprietary LMMs}, including GPT-5~\cite{gpt5} and Gemini-3-Pro~\cite{gemini3pro}.
\end{itemize}

For \textbf{visual quality comparison}, we further compare QualiRAG with representative VQA methods specifically designed for quality scoring. For images, these include MUSIQ~\cite{ke2021musiq}, TRES~\cite{golestaneh2022no}, CLIP-IQA+~\cite{wang2023exploring}, LIQE~\cite{zhang2023blind}, TOPIQ~\cite{chen2024topiq} and Q-Align~\cite{wu2023q}; for videos, we consider FAST-VQA~\cite{wu2022fast}, DOVER~\cite{wu2023exploring}, COVER~\cite{he2024cover} and MinimalisticVQA~\cite{sun2024analysis}. For VQA methods trained to output scalar quality scores, we compute score differences between paired images or videos and evaluate pairwise accuracy. For question-answering-based methods, we report their classification accuracy on the binary-choice questions: \texttt{``Which video/image has better visual quality?''} with candidate answers: \texttt{(A) The first video/image} and \texttt{(B) The second video/image}.
 
\vspace{0.1cm}

\noindent\textbf{Implementation Details.} 
The $\text{LMM}_{\text{main}}$ and $\text{LMM}_{\text{aux}}$ are instantiated by InternVL3-9B-Instruct and Qwen3-VL-8B-Instruct respectively. As analyzed in Section~\ref{performance_analysis}, InternVL3-9B-Instruct exhibits relatively limited performance on visual quality understanding tasks, thereby providing a suitable testbed to verify that the observed performance gains are attributable to QualiRAG rather than the intrinsic capability of the backbone. In contrast, Qwen3-VL-8B-Instruct demonstrates strong capability in perceptual quality description, making it well suited for constructing auxiliary quality knowledge. For video inputs, we sample $1$ fps as the visual input. Visual inputs processed by Qwen3-VL-8B-Instruct are kept at their original resolution to support fine-grained auxiliary knowledge construction, whereas inputs to InternVL3-9B-Instruct are resized to $448\times448$ following its default input configuration. The hyperparameters $\tau$ in Eq.~\eqref{eq:sim} and $n_l$ in Eq.~\eqref{local} are set to $0.25$ and $4$, respectively.
\subsection{Performance Analysis}
\label{performance_analysis}
\noindent\textbf{Visual Quality Understanding.}
The results on the {\tt dev} subsets of Q-Bench and Q-Bench-Video are reported in Table~\ref{tab:q-bench-dev} and Table~\ref{tab:q-bench-video-dev}, respectively. On Q-Bench, QualiRAG achieves an overall accuracy of $81.74\%$, \textbf{outperforming all open-source general LMMs, VQA-finetuned LMMs, and even surpassing strong proprietary models} GPT-5 and Gemini-3-Pro. Similarly, on Q-Bench-Video, QualiRAG attains an overall accuracy of $61.66\%$, \textbf{outperforming all open-source general LMMs and VQA-finetuned LMMs}, including Q-route, an agentic framework that leverages a diverse set of VQA expert models with GPT-4o~\cite{achiam2023gpt} as its backbone. Moreover, QualiRAG surpasses GPT-5 and trails Gemini-3-Pro by only $1.36\%$.  These results demonstrate that retrieval-augmented generation with structured query decomposition and relevance-aware retrieval over complementary global and local quality evidence effectively improves visual quality understanding, without relying on task-specific instruction tuning or VQA expert models.

\vspace{0.1cm}

\noindent\textbf{Visual Quality Comparison.}
The performance on image and video quality comparison tasks is reported in Table~\ref{tab:image_quality} and Table~\ref{tab:video_quality}, respectively. For image quality comparison, \textbf{QualiRAG achieves the best overall performance among all evaluated methods}. Most competing approaches perform well on LIVE-C, which serves as an in-domain benchmark for methods trained on the KonIQ dataset~\cite{hosu2020koniq}, but exhibit noticeable performance degradation on AGIQA, containing AI-generated images, and on PIPAL, focusing on processing-induced distortions. In contrast, QualiRAG maintains consistently strong performance across all image benchmarks. This indicates that the proposed training-free RAG framework provides more robust image quality assessment under out-of-domain settings. For video quality comparison, QualiRAG also demonstrates strong competitiveness. \textbf{It substantially outperforms open-source general-purpose LMMs and achieves performance comparable to DNN-based VQA models} trained on the LSVQ dataset~\cite{ying2021patch}. While LMM-based methods fine-tuned on LSVQ (\textit{e.g.}, VQAThinker) attain higher overall accuracy, QualiRAG delivers more favorable performance on out-of-domain benchmarks such as VDPVE and LIVE-HFR. This observation further highlights the robustness of QualiRAG for visual quality comparison, a task that directly reflects visual quality rating capability.

\begin{table}[t]\small
    \centering
    \renewcommand\arraystretch{1.1}
    \renewcommand\tabcolsep{18pt}
    \belowrulesep=0pt\aboverulesep=0pt
    \caption{Ablation study of the contributions of four complementary knowledge databases to overall performance on the {\tt dev} subset of Q-Bench-Video. $^*$ indicates that $\text{LMM}_\text{aux}$ is replaced with the VQA-finetuned model VQA$^2$.}
    \vspace{-8pt}
    \resizebox{\linewidth}{!}{\begin{tabular}{cccc|c}
    \toprule

        {\textbf{$\mathcal{K}_{\text{meta}}$}} & {\textbf{$\mathcal{K}_{\text{loc}}$}} & \textbf{$\mathcal{K}_{\text{globalQ}}$} & \textbf{$\mathcal{K}_{\text{localQ}}$} & {Overall $\uparrow$} \\ \hline
         & & & & 47.61\% \\
         \ding{52} & & & & 49.22\% \\
         \ding{52} & \ding{52} & & & 51.19\% \\
         \ding{52} & \ding{52} & \ding{52} & & 55.84\% \\
         \ding{52} & \ding{52} & \ding{52} & \ding{52} & 61.66\% \\
         \ding{52} & \ding{52} & \ding{52}$^*$ & \ding{52}$^*$ & 64.12\%\\
    
       \bottomrule
    \end{tabular}}
    \vspace{-5pt}
    \label{tab:ablation}
\end{table}

\subsection{Ablation Study}

We perform ablation studies on the four complementary quality-related knowledge databases and on different backbone LMMs used in the Answer Generator. All ablation studies are performed on Q-Bench-Video, as it poses a more challenging setting for visual quality understanding.

\vspace{0.1cm}
\noindent\textbf{Knowledge Databases.}
The knowledge databases $\mathcal{K}$ constitute a core component of QualiRAG for enhancing the visual quality understanding capability. To quantify the contribution of each auxiliary database, we progressively enable the four knowledge sources and additionally replace the auxiliary model $\mathrm{LMM}_{\text{aux}}$ from Qwen3-VL-8B-Instruct with the VQA$^2$ expert to derive more VQA-specific knowledge. This procedure yields a clear monotonic improvement in performance.

Specifically, introducing the metadata database $\mathcal{K}_{\text{meta}}$ yields a $1.61\%$ gain, indicating that low-level technical metadata provides useful contextual cues for quality reasoning. Adding the subject localization database $\mathcal{K}_{\text{loc}}$ further improves performance by $1.97\%$, demonstrating that explicit spatial grounding helps the model focus on question-relevant regions. Incorporating the global quality summary database $\mathcal{K}_{\text{globalQ}}$ results in a more substantial improvement of $4.65\%$, highlighting the importance of holistic perceptual quality descriptions. The local quality description database $\mathcal{K}_{\text{localQ}}$ delivers the largest gain of $5.82\%$, underscoring the critical role of fine-grained, question-aligned local quality evidence in visual quality understanding. When $\mathrm{LMM}_{\text{aux}}$ is further replaced with the VQA$^2$ expert model, the overall accuracy reaches $64.12\%$, representing the best performance among all variants and even surpassing Gemini-3-Pro.

\begin{figure}[t]
\centering
\vspace{-0.2cm}
\centerline{\epsfig{figure=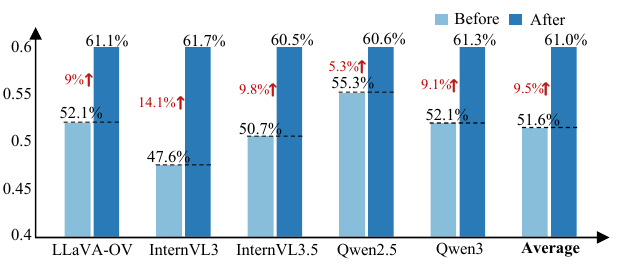,width=9cm}}
\vspace{-0.2cm}
\caption{Performance of LLaVA-OneVision-7B, InternVL3-9B-Instruct, InternVL3.5-8B-Instruct, Qwen2.5-VL-7B-Instruct, and Qwen3-VL-8B-Instruct on the {\tt dev} subset of Q-Bench-Video, before and after applying QualiRAG.} 
\label{fig:backbone_ablation}
\vspace{-0.4cm}
\end{figure}

\vspace{0.1cm}
\noindent\textbf{Backbone-Agnostic Performance.}
We investigate the impact of the backbone $\text{LMM}_\text{main}$ on QualiRAG by replacing the default backbone with a diverse set of open-source general-purpose LMMs. As shown in Figure~\ref{fig:backbone_ablation}, without QualiRAG, the evaluated backbones exhibit notable performance differences, and even the strongest model, Qwen2.5-VL, achieves only moderate accuracy. After incorporating QualiRAG, all backbones achieve consistent performance improvements and converge to a similar level of approximately $61\%$, with variations within $1\%$. This convergence indicates that the effectiveness of QualiRAG is largely backbone-agnostic, suggesting that the observed improvements primarily stem from the quality-related knowledge provided by the framework rather than the intrinsic capabilities of individual backbones. These results demonstrate that \textsc{QualiRAG} is robust to backbone choices and can be seamlessly applied to a wide range of LMM architectures.

\section{Discussion}

Visual quality perception is a fundamental capability of the human visual system, yet it has received limited attention in the development of large multimodal models. In this work, the proposed QualiRAG framework demonstrates that grounding inference on quality-related evidence can substantially enhance the visual quality perception ability of LMMs. Compared to training-based paradigms, QualiRAG offers significant advantages in \textbf{scalability}. \textit{First}, QualiRAG is entirely training-free, enabling seamless integration with updated or more powerful vision-language backbones. \textit{Second}, the auxiliary knowledge databases can be dynamically extended according to specific evaluation requirements; for instance, when assessing AI-generated content, domain-specific artifact detectors~\cite{huang2024vbench} can be incorporated to explicitly capture AIGC-related distortions. \textit{Third}, QualiRAG is complementary to existing VQA models; for example, replacing the quality description generators with VQA$^2$ improves performance on Q-Bench-Video beyond Gemini-3-Pro, highlighting their synergistic potential. Overall, QualiRAG provides a flexible and extensible approach to visual quality assessment that can adapt to a broad range of real-world scenarios.

\section{Conclusion}

We propose QualiRAG, a training-free retrieval-augmented generation framework for visual quality understanding.
Instead of relying on task-specific fine-tuning based on large-scale quality instruction data, QualiRAG improves visual quality perception ability by grounding inference on retrieved quality-related evidence.
By decomposing quality-related questions into retrieval-oriented queries and integrating complementary knowledge sources at different granularity levels, QualiRAG enables effective and interpretable visual quality analysis. Experimental results validate the effectiveness of QualiRAG for both visual quality understanding and quality comparison.

\bibliographystyle{named}
\bibliography{ijcai26}

\appendix

\twocolumn[
\begin{center}
    \LARGE \textbf{QualiRAG: Retrieval-Augmented Generation for Visual Quality Understanding} \\
    Appendix
\end{center}]

\begin{table*}[t]\small
    \centering
    \renewcommand\arraystretch{1.1}
    \renewcommand\tabcolsep{7pt}
    \belowrulesep=0pt\aboverulesep=0pt
    \caption{An overview of our testing Benchmarks.}
    \vspace{-8pt}
    \resizebox{\linewidth}{!}{\begin{tabular}{lccccccc}
    \toprule
        \textbf{Benchmark} & \textbf{Year} & \textbf{\# of Data} & \textbf{\# of Scenes} & \textbf{Resolution} & \textbf{Duration} & \textbf{Frame Rate} & \textbf{Distortion Type}  \\ 
        \hline
        \rowcolor{gray!10}
        \multicolumn{8}{l}{\textit{Benchmarks for Visual Quality Understanding}} \\ 
        Q-Bench & 2023 & 2,990 & 2,990 & Diverse & - & - & Diverse \\
        Q-Bench-Video & 2025 & 1,800 & 1,800 & Diverse & Diverse & Diverse & Diverse \\  
        \hline
        \rowcolor{gray!10}
        \multicolumn{8}{l}{\textit{Benchmarks for In-domain Visual Quality Rating}} \\ 
        LIVE-C & 2016 & 1,169 & 1,169 & 500x500 & - & - &  In-the-wild\\ 
        KoNViD-1k & 2017 & 1,200 & 1,200 & 540p & 8 & 24, 25, 30 & In-the-wild \\
        \hline
        \rowcolor{gray!10}
        \multicolumn{8}{l}{\textit{Benchmarks for Out-of-domain Visual Quality Rating}} \\
        PIPAL & 2020 & 29,000 & 250 & 288x288 & - & - & synthetic \\
        AGIQA & 2023 & 2,982 & 2,982 & 512x512 & - & - & AIGC \\
        LIVE-YT-HFR & 2021 & 480 & 16 & 1080p & 6-10 & 24, 30, 60, 82, 98, 120 &  Frame rate, VP9 compression \\
        VDPVE & 2023 & 1,211 & 79 & Diverse & 8-15 & 24, 25, 30 & Enhancement \\
       \bottomrule
    \end{tabular}}
    \vspace{-5pt}
    \label{tab:dataset_summary}
\end{table*}

\vspace{0.4cm}

\section{More Details of Our Testing Benchmarks}

Table~\ref{tab:dataset_summary} provides an overview of our testing benchmarks, which encompass diverse content types, resolutions, durations, frame rates, and distortion types. In the following, we provide a detailed description of each benchmark.

\begin{itemize}  
    \item \textbf{Q-Bench~\cite{wu2023qbench}}: Q-Bench is a benchmark designed to systematically evaluate the low-level visual perception capabilities of large multimodal models (LMMs) across diverse attributes and conditions. It introduces the LLVisionQA dataset, which contains 2{,}990 images paired with human-authored low-level visual questions, covering a wide range of image sources, including in-the-wild photography, AI-generated content (AIGC), and artificially distorted images. LLVisionQA comprises three types of questions—Yes-or-No, What, and How—and organizes low-level visual concerns into four quadrants along two axes: distortion versus other low-level attributes, and global perception versus local content-related perception. Together, these design choices provide a holistic, diverse, and balanced benchmark for assessing LMMs’ low-level visual perception abilities.
    
    \item \textbf{Q-Bench-Video~\cite{zhang2025q}:} Q-Bench-Video is a benchmark designed to systematically evaluate the video quality understanding capabilities of large multimodal models (LMMs). It covers a wide spectrum of video content, including natural scenes, AIGC, and computer graphics. To ensure a balanced distribution of perceptual quality, videos are sampled from multiple sources with available subjective annotations. The benchmark includes both multiple-choice and open-ended questions, enabling the evaluation of model performance under diverse quality assessment settings. Overall, Q-Bench-Video consists of 1{,}800 videos and 2{,}378 annotated question–answer pairs for validation, providing a comprehensive testbed for assessing video quality understanding in LMMs.
    
    \item \textbf{LIVE-C~\cite{ghadiyaram2015massive}:} LIVE-C (LIVE In the Wild Image Quality Challenge Database) is a large-scale visual quality assessment (VQA) benchmark designed to address the limitations of traditional datasets built under controlled laboratory conditions. Unlike synthetic-distortion datasets, LIVE-C contains authentically distorted images captured using a wide range of real-world mobile devices, reflecting complex mixtures of distortions commonly encountered in practice. The dataset is constructed via a large-scale crowdsourcing study, comprising over 350{,}000 subjective opinion scores collected from more than 8{,}100 human observers on 1{,}169 images. 

    \item \textbf{PIPAL~\cite{jinjin2020pipal}:} PIPAL (Perceptual Image Processing Algorithms) is a large-scale VQA dataset designed to evaluate both VQA methods and modern image restoration algorithms, particularly those based on Generative Adversarial Networks (GANs). The dataset contains 29{,}000 distorted images generated from 40 distortion types and 116 distortion levels. PIPAL is annotated with over 1.13 million human judgments collected using an Elo-based~\cite{elo1978rating} subjective evaluation system, providing reliable perceptual quality scores. In this work, we evaluate our model on the validation subset containing 1,000 images.
    
    \item \textbf{AGIQA~\cite{li2023agiqa}:} AGIQA is a large-scale subjective image quality assessment dataset specifically designed for AI-generated images (AGIs). It contains 2{,}982 images generated by six representative text-to-image models, covering diverse generation paradigms including GAN-, autoregressive-, and diffusion-based models. The dataset is constructed by carefully varying prompts and internal generation parameters to capture a wide range of perceptual quality variations. 

    \item \textbf{KoNViD-1k~\cite{hosu2017konstanz}:} KoNViD-1k is an authentic video quality assessment dataset comprising 1{,}200 unique test videos that exhibit a wide range of real-world distortions. All videos are sampled from the YFCC100M dataset~\cite{thomee2016yfcc100m} using a feature-based selection strategy that accounts for blur, colorfulness, contrast, spatial and temporal information, as well as the no-reference image quality metric NIQE~\cite{mittal2012making}. Each video is clipped from the original source content and resized to 540p with a landscape aspect ratio. The videos have frame rates of 24, 25, or 30 fps, and each clip has a fixed duration of 8 seconds.

    \item \textbf{LIVE-YT-HFR~\cite{madhusudana2021subjective}:} LIVE-YT-HFR is a video quality assessment dataset designed to investigate the perceptual effects of frame rate variation and compression on video quality. The dataset comprises 16 source sequences and 480 distorted videos, generated by applying six different frame rates in conjunction with five levels of VP9 compression, including one lossless setting and four CRF-based compression levels. Among the source sequences, 11 are drawn from the BVI-HFR dataset~\cite{mackin2018study}, featuring a resolution of 1920$\times$1080 and a fixed duration of 10 seconds. The remaining five sequences consist of high-motion sports content captured by Fox Media Group, provided at a resolution of 3840$\times$2160, with durations ranging from 6 to 8 seconds.

    \item \textbf{VDPVE~\cite{gao2023vdpve}:} VDPVE is a video quality assessment VQA dataset developed for perceptual video enhancement. It comprises a total of 1{,}211 enhanced videos, organized into three subsets: (1) 600 videos with color, brightness, and contrast enhancements; (2) 310 videos with deblurring enhancements; and (3) 301 videos with deshaking enhancements. In this work, we evaluate our model on an open-sourced subset of 839 videos from the training split.

\end{itemize}

\section{More Details of Our QualiRAG Model}


\subsection{Query Organizer}
The decoupling instruction $\mathcal{P}_{\text{dec}}$ used in QualiRAG for transforming raw questions into structured retrieval requests is defined as follows:

\begin{tcolorbox}[
    title={Decoupling Instruction $\mathcal{P}_{\text{dec}}$},
    colback=gray!5,
    colframe=black,
    fonttitle=\bfseries,
    breakable
]
You will analyze the following visual-question and decompose it into a structured quality-analysis schema. \\

Your job is NOT to answer the question. \\

Instead, extract: \\
1. The subject referenced in the question. (Identify the concrete, visually identifiable entity mentioned in the question (e.g., child, person, road, building, mountain, object). Abstract terms such as video, image quality, or distortion should not be considered valid subjects. If no valid subject exists, set it to "none".) \\
2. The relevant quality dimension(s) such as clarity, sharpness, blur, noise, blockiness, temporal, etc. \\
3. The main scope category: spatial or temporal. \\
4. Determine the most relevant visual area or time segment the LMM should attend to when assessing the specified subject and quality dimension. \\

Return your output in JSON format: \\

\{ \\
``subject'': ``...'', \\
``dimension'': ``...'',\\
``scope'': ``...'',\\
``focus'': ``...''\\
\} \\

Here is the question:
$[\texttt{question}]$
\end{tcolorbox}

\subsection{Source Augmenter}
\noindent\textbf{Metadata Database.}
For video inputs, the spatial resolution $r$, frame rate $f$, video duration $d$, and average bitrate $b_{\text{avg}}$ are directly extracted using \texttt{ffprobe}~\cite{ffprobe} from the video stream metadata. These attributes describe the basic acquisition and encoding configuration of the video and are obtained without any additional processing.

In addition to these static properties, we introduce a temporal bitrate variation indicator $b_{\text{trend}}$ to capture coarse-grained bitrate dynamics over time. To compute this indicator, we estimate the average video bitrate over two temporal segments of the video: the head segment and the tail segment. Specifically, we sample the first 10\% and the last 10\% of the video duration, respectively, and compute the mean bitrate over all frames within each segment using frame-level bitrate statistics provided by \texttt{ffprobe}. Let $\bar{b}_{\text{head}}$ and $\bar{b}_{\text{tail}}$ denote the estimated average bitrates of the head and tail segments. The relative bitrate change ratio is then defined as
\[
\Delta = \frac{\bar{b}_{\text{tail}} - \bar{b}_{\text{head}}}{\max(\bar{b}_{\text{head}}, 1)}.
\]
Based on this ratio and the overall average bitrate $b_{\text{avg}}$, we categorize the temporal bitrate trend $b_{\text{trend}}$ into four discrete types:
\begin{itemize}
    \item \textit{Increasing}, if $\Delta > 0.1$;
    \item \textit{Decreasing}, if $\Delta < -0.1$;
    \item \textit{Constant high}, if $|\Delta| \leq 0.1$ and $b_{\text{avg}}$ exceeds a predefined threshold $\tau_b = 3$ Mbps;
    \item \textit{Constant low}, if $|\Delta| \leq 0.1$ and $b_{\text{avg}} \leq \tau_b$.
\end{itemize}

\noindent\textbf{Global Quality Summary Database.} The detailed global quality query $Q_{\text{global}}$ is given as follows:

\begin{tcolorbox}[
    title={Global Quality Query $Q_{\text{global}}$},
    colback=gray!5,
    colframe=black,
    fonttitle=\bfseries,
    breakable
]
For the given image or video, please provide an overall perceptual quality description in terms of structure and texture preservation, color and luminance reproduction, noise, contrast, sharpness, motion, artifacts, stutter, and other low-level distortions, and explain the reasons for your assessment.

\end{tcolorbox}

\noindent\textbf{Local Quality Description Database.}
Given a structured request $\mathbf{R}$, we construct the local quality query $Q_{\text{local}}$ according to the number of visual inputs. For a single visual input, the query takes the form: \texttt{``Describe the $\mathbf{R}$[dimension] of the $\mathbf{R}$[subject] in the image/video.''} For multiple visual inputs, the query is adapted to: \texttt{``Compare the $\mathbf{R}$[dimension] of the $\mathbf{R}$[subject] across all images/videos.''}

To improve robustness and mitigate stochastic variability in local quality description generation, we query the LMM multiple times to obtain $n_l$ independent local quality descriptions for each question. During generation, we adopt temperature-based sampling with $\texttt{temperature}=1.0$ and nucleus sampling ($\texttt{top\_p}=0.95$) to balance descriptive diversity and generation stability. Hard top-$k$ truncation is disabled by setting $\texttt{top\_k}=0$, thereby avoiding excessive pruning of informative quality-related tokens. The prompt used to aggregate consistent information across these descriptions is as follows:

\begin{tcolorbox}[
    title={Aggregation Information Prompt},
    colback=gray!5,
    colframe=black,
    fonttitle=\bfseries,
    breakable,
    listing only,
    listing options={
        basicstyle=\ttfamily\footnotesize,
        breaklines=true,
        breakatwhitespace=true
    }
]

You are given $n_l$ different descriptions of the visual quality of the SAME visual input.\\

Your task is to produce ONE final visual quality description that reflects the shared, consistent judgment across these descriptions. \\

Important rules:

- Do NOT mention that multiple answers were given.

- Do NOT use phrases like ``most responses agree'', ``overall'', or ``in summary''.

- Write as if you directly observed the image/video yourself. \\

Here are the descriptions:

$\{\texttt{numbered\_descriptions}\}$ \\

Write ONE coherent visual quality description.

\end{tcolorbox}

\subsection{Source Selector}

We adopt \textbf{Contriever} framework~\cite{izacard2021unsupervised} as the text encoder to map both the quality-related question $Q$ and each candidate knowledge sentence $d$ into a shared embedding space. Given an input text $x$, its vector representation is obtained by mean pooling over the last-layer hidden states of the encoder:
\begin{equation}
\mathbf{e}(x)=\mathrm{MeanPool}\big(\mathrm{Contriever}(x)\big)\in\mathbb{R}^{m}.
\end{equation}

Both the question embedding and the candidate knowledge sentence embeddings are $\ell_2$-normalized to compute inner-product similarity:
\begin{equation}
\hat{\mathbf{e}}(x)=\frac{\mathbf{e}(x)}{\|\mathbf{e}(x)\|_2}, \quad
\mathrm{sim}(Q,d)=\hat{\mathbf{e}}(Q)^{\top}\hat{\mathbf{e}}(d).
\end{equation}

All knowledge sentence embeddings are indexed using \textbf{FAISS} library~\cite{johnson2019billion} to enable efficient approximate nearest-neighbor search. Rather than selecting a fixed top-$k$, we perform range search with a predefined similarity threshold $\tau$, retaining only knowledge sentences whose relevance scores exceed $\tau$:
\begin{equation}
\mathcal{K}^{*}(Q)=
\left\{ d \in K \;\middle|\; \mathrm{sim}(Q,d)\ge\tau \right\}.
\end{equation}

The retrieved knowledge sentences are then concatenated into a paragraph, which is subsequently provided to the Answer Generator for knowledge-grounded quality reasoning.

\subsection{Answer Generator}
The prompt used in the final Answer Generator is as follows.

For a single visual input:

\begin{tcolorbox}[
    title={Answer Generator Prompt},
    colback=gray!5,
    colframe=black,
    fonttitle=\bfseries,
    breakable,
    listing only,
    listing options={
        basicstyle=\ttfamily\footnotesize,
        breaklines=true,
        breakatwhitespace=true
    }
]

Now you will receive an image/video: \texttt{[visual$\_$token]}. \\

Here is the metadata for the image/video: $[\mathcal{K}^*_{\text{meta}}]$. 

Here is the localization description of objects in the image/video for reference: $[\mathcal{K}_{\text{loc}}]$ or \texttt{NULL}.

Here is the global quality summary of the image/video for reference: $[\mathcal{K}^*_{\text{globalQ}}]$.

Here is the local quality description of the image/video for reference: $[\mathcal{K}^*_{\text{localQ}}]$. \\

You are performing a visual quality understanding task. Here is the question:

\texttt{[question]} \\

Please combine what you observe in the image/video with the reference descriptions to make your judgment.
In your response, act as if your conclusion is drawn solely from the visual content itself, rather than from the reference descriptions. \\

Pay special attention to $\mathbf{R}[\texttt{focus}]$, as well as both global and local distortions in the image/video, including blur, noise, compression artifacts, color inconsistencies, and motion instability.
Your evaluation should reflect how these distortions affect the overall perceptual quality. \\

First, output your reasoning process enclosed in $\texttt{<think>}$ and $\texttt{</think>}$ tags.
Then, output the selected option letter (if applicable) directly in $\texttt{<answer>}$ and $\texttt{</answer>}$ tags.

\end{tcolorbox}

For a pair of visual inputs:

\begin{tcolorbox}[
    title={Answer Generator Prompt},
    colback=gray!5,
    colframe=black,
    fonttitle=\bfseries,
    breakable,
    listing only,
    listing options={
        basicstyle=\ttfamily\footnotesize,
        breaklines=true,
        breakatwhitespace=true
    }
]

Now you will receive the first image/video: \texttt{[visual$\_$token]$_1$}, the second image/video: \texttt{[visual$\_$token]$_2$}. \\

Here is the metadata about the first image/video: $[\mathcal{K}^*_{\text{meta}}]_1$.

Here is the metadata about the second image/video: $[\mathcal{K}^*_{\text{meta}}]_2$.

This is the localization description of objects in the the first image/video for reference: $[\mathcal{K}^*_{\text{loc}}]_1$ or \texttt{NULL}.

This is the localization description of objects in the second image/video for reference: $[\mathcal{K}^*_{\text{loc}}]_2$ or \texttt{NULL}.

This is the global quality summary of the first image/video for reference: $[\mathcal{K}^*_{\text{globalQ}}]_1$.

This is the global quality summary of the second image/video for reference: $[\mathcal{K}^*_{\text{globalQ}}]_2$.

This is the local quality description of the two images/videos for reference: $[\mathcal{K}^*_{\text{localQ}}]$. \\

You are performing a visual quality understanding task. Here is the question:

\texttt{[question]} \\

Please combine what you observe in the image/video with the reference descriptions to make your judgment.
In your response, act as if your conclusion is drawn solely from the visual content itself, rather than from the reference descriptions. \\

Pay special attention to $\mathbf{R}[\texttt{focus}]$, as well as both global and local distortions in the image/video, including blur, noise, compression artifacts, color inconsistencies, and motion instability.
Your evaluation should reflect how these distortions affect the overall perceptual quality. \\

First, output your reasoning process enclosed in $\texttt{<think>}$ and $\texttt{</think>}$ tags.
Then, output the selected option letter (if applicable) directly in $\texttt{<answer>}$ and $\texttt{</answer>}$ tags.

\end{tcolorbox}


\section{More Visualizations of Performance}
\begin{figure*}[t]
\centering
\centerline{\epsfig{figure=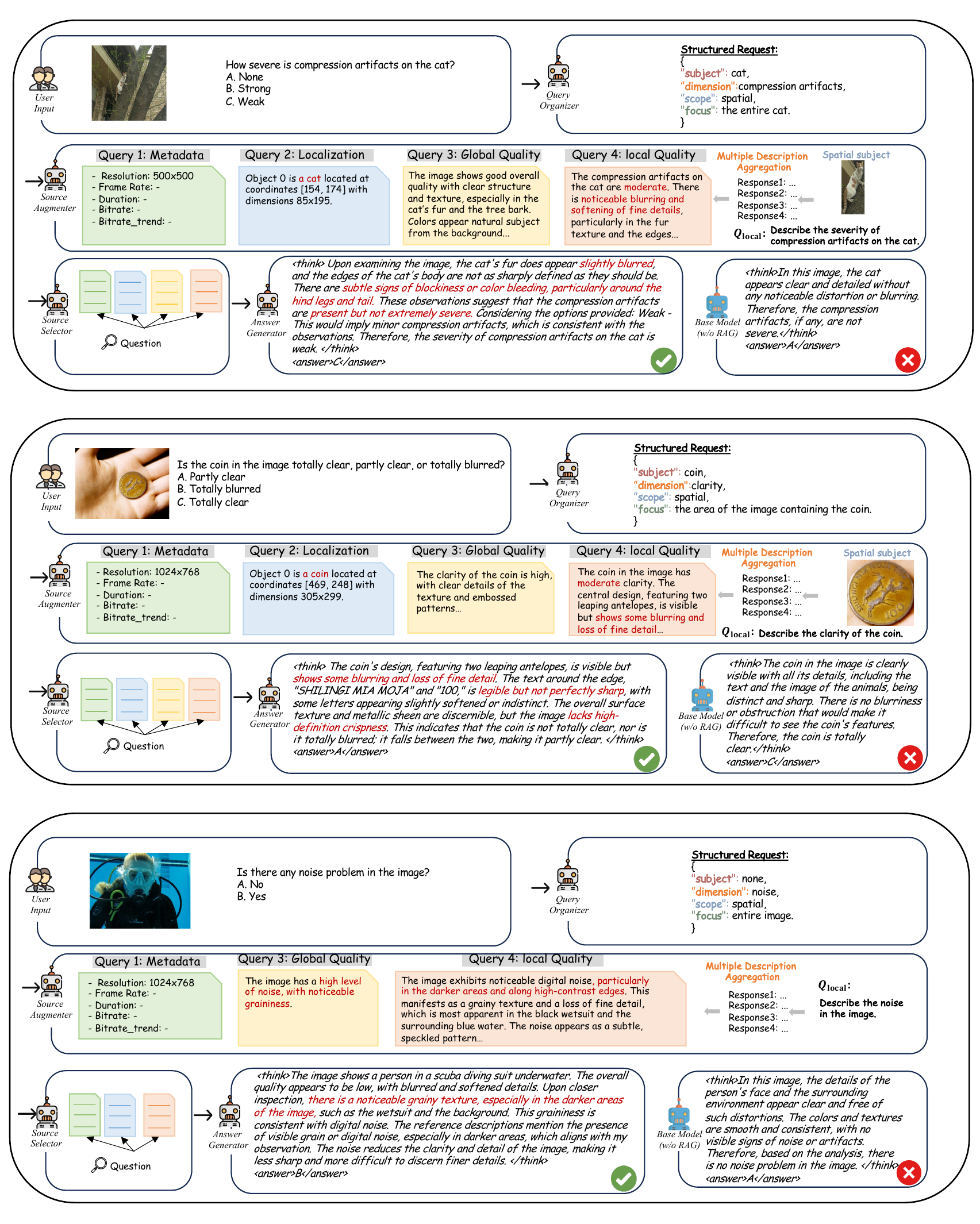,width=18cm}}
\caption{Examples of image quality understanding on Q-Bench.}
\label{fig:supp_q-bench}
\end{figure*}

\begin{figure*}[t]
\centering
\centerline{\epsfig{figure=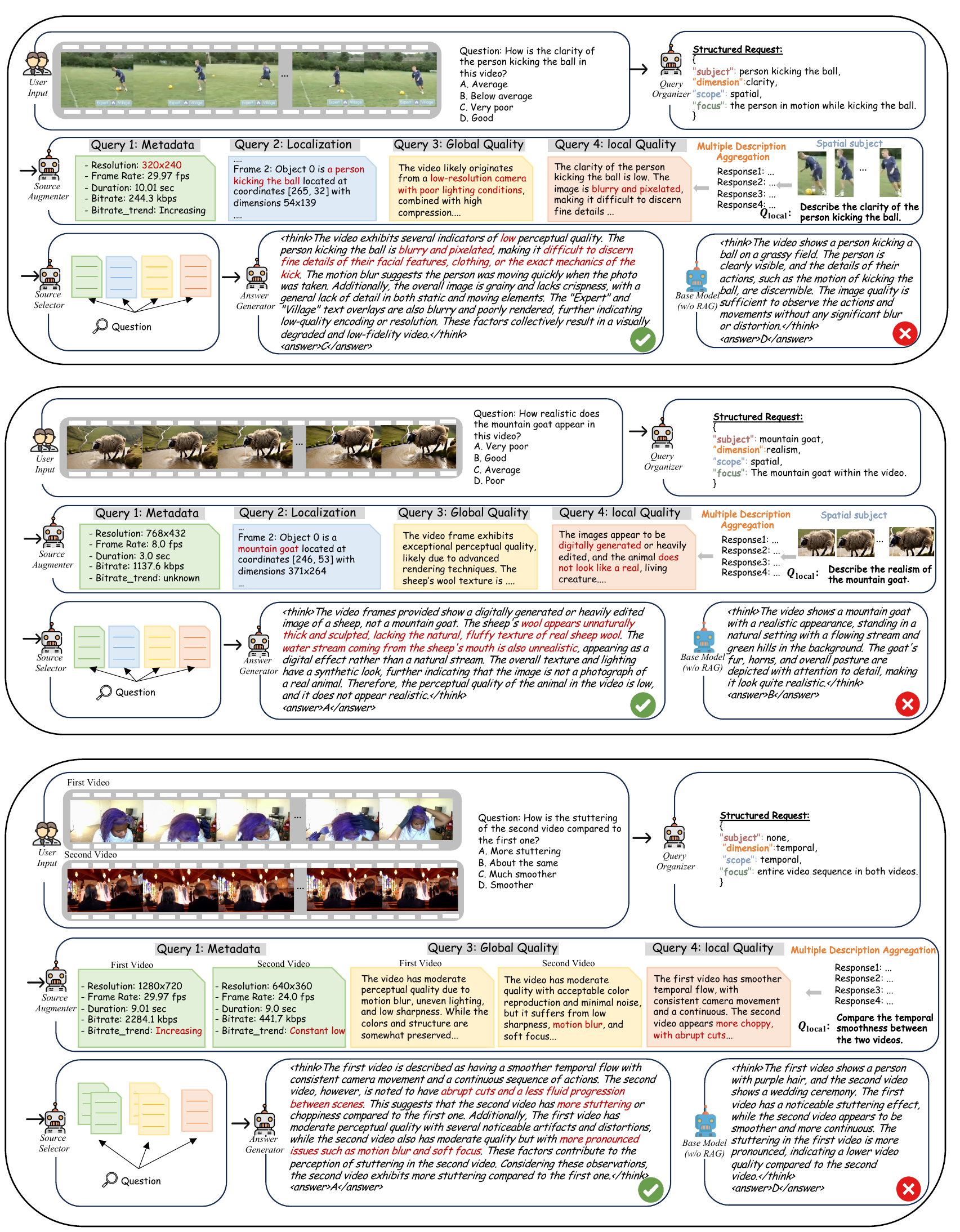,width=18cm}}
\caption{Examples of video quality understanding on Q-Bench-Video.}
\label{fig:supp_q-bench-video}
\end{figure*}

Figures~\ref{fig:supp_q-bench} and~\ref{fig:supp_q-bench-video} show additional examples of the visual quality understanding ability of our QualiRAG.

\end{document}